\documentclass[runningheads]{llncs}
\usepackage[style=numeric,backend=biber]{biblatex}
\usepackage[T1]{fontenc}
\usepackage{graphicx}
\begin{document}
\title{Enhancing Social Robots through Resilient AI}
%
%
\author{Domenico Palmisano\inst{1}\and
Giuseppe Palestra\inst{1} \and
Berardina Nadja De Carolis\inst{2}}
\authorrunning{D. Palmisano et al.}
%
\institute{Hero srl, Apulia, Italy 
\email{info@herovision.it} \and
University of Bari ``Aldo Moro", Italy
\email{berardina.decarolis@uniba.it}}
%
\maketitle              
\begin{abstract}
As artificial intelligence continues to advance and becomes more integrated into sensitive areas like healthcare, education, and everyday life, it's crucial for these systems to be both resilient and robust. This paper shows how resilience is a fundamental characteristic of social robots, which, through it, ensure trust in the robot itself—an essential element especially when operating in contexts with elderly people, who often have low trust in these systems.
Resilience is therefore the ability to operate under adverse or stressful conditions, even when degraded or weakened, while maintaining essential operational capabilities.

\end{abstract}

\section{Introduction}
The evolution of artificial intelligence systems and their increasing use in complex and sensitive situations, such as healthcare, education, and daily life, require AI systems to become increasingly resilient and robust. The concept of resilience within AI systems is seen as the system’s ability to make decisions and operate in critical or suboptimal situations, for example, with partial or imperfect inputs, as well as its ability to adapt to the surrounding environment.
The concept of resilience also plays a fundamental role in social robotics, since this field aims to develop robots capable of interacting with humans in a natural and emotionally aware manner, adapting to the mental models and behavior of the users. It is therefore essential to ensure resilience in social robots as well, to enable better and more adapted human-robot interaction.
This paper analyzes the general concept of resilient AI, followed by a focus on the application of resilient approaches in typical social robotic tasks adapt to the user such as facial emotion recognition (FER) and natural language processing (NLP).
The paper examines the techniques employed in these two fundamental phases of the development of resilient AI systems, with a particular focus on FER and NLP.

Both of these phases must take the concept of resilience into account: in the first phase, resilience is introduced primarily through a Data Augmentation process, with the aim of adding to the data situations typical of erroneous or incomplete inputs; in the second phase, resilience is implemented through the study of more stable and reliable models.
This study highlights that in order for an artificial intelligence system to be truly resilient, its development must follow a targeted design process capable of integrating resilience-oriented principles and strategies from the earliest stages.
The study presents 20 scientific articles published between 2021 and 2025. The main objective is to identify the most effective strategies for increasing the reliability and adaptability of AI in dynamic, unpredictable contexts affected by noise and data ambiguity.



\section{The RAISE project}
The RAISE (Resilient AI Systems for hEalth) project aims at the development of a resilient artificial intelligence system designed to support the elderly population (over 65 years) in domestic and care settings. This paper represents the first step of the project: a critical review of resilient AI algorithms. This critical review is necessary to ensure that such systems are specifically designed to meet the needs of safety, well-being, and socialization, both in general and, more specifically, in the fields of FER and NLP.
Thanks to the project, the integration of resilient AI algorithms in social robots opens up new possibilities for the development of more sophisticated, autonomous, and intelligent social robotic systems in complex environments.




\section{Related work}



Resilience is a widely used but conceptually diverse term, and a careful unpacking of its meanings is necessary when positioning research in social robotics and resilient AI. Different disciplines frame resilience through distinct lenses: ecology emphasizes the capacity of a system to absorb disturbances and remain within a stable regime \cite{holling1973resilience}; psychology stresses processes of recovery and positive adaptation at the individual level \cite{southwick2014resilience}; engineering focuses on a system’s functional abilities to anticipate, detect, respond to, and learn from disruptions \cite{hollnagel2011scope}. 

From an ecological standpoint, resilience originally referred to the ability of ecosystems to withstand perturbations without undergoing a regime shift: the concern is with persistence, thresholds, and the magnitude of disturbance a system can tolerate while retaining core structure and function. In contrast, psychological and developmental literatures treat resilience as a set of adaptive processes and outcomes — trajectories of adjustment, the role of protective mechanisms (e.g., social support, coping strategies), and the dynamic interplay between risk and resources over time. Engineering and safety-science accounts recast resilience in functional terms: what can the system do when events deviate from expectations? Resilience engineering foregrounds capabilities such as anticipation, continuous monitoring, timely response, and organizational learning as means to sustain critical functions under unexpected stressors.


For the purposes of this paper we adopt an explicit, multidimensional operationalization of resilience that maps across the literatures encountered in social robotics and resilient AI: (i) absorptive capacity — the ability to tolerate disturbance and continue functioning (persistence) \cite{holling1973resilience}; (ii) adaptive capacity — the ability to reorganize or reconfigure so that core functions continue despite change \cite{masten2001ordinary}; and (iii) transformative capacity — the ability to fundamentally change structure or behaviour, enabling qualitatively new functions or regimes \cite{hollnagel2021resilience}. 


\subsection{Resilient AI in social robotics}

Artificial intelligence can improve resilience in social robotics for various domains, including healthcare, education, and emergency management. In a recent paper \cite{rane2024artificial}, AI has been shown to improve the ability of complex systems (in multiple sectors) to prevent, manage, and respond to critical events through continuous learning models and predictive and adaptive algorithms. In the study, resilience is treated as a dynamic property of a complex system, enabled by AI technologies that learn from data and adapt the system to the user or the environment.

Artificial intelligence architectures need to be capable of autonomously evolving and adapting to changing operating conditions. In order to face up to this issue, AI must overcome the limitation of static structures and use self-configuration and self-optimization mechanisms. In \cite{kunduself}, a new AI architecture based on this concept has been proposed. Three main components have been defined: adaptive learning, architectural modularity, and intelligent management of computational resources. These components aims to create a resilient AI system capable of autonomously recognizing when a change is needed and implementing it.
A social robot to be resilient requires flexible and adaptable AI algorithms. As humans or animals possess learning and resilience capabilities in the study \cite{saxena2024ai} concepts such as episodic memory, selective attention, synaptic plasticity, and reward are treated as possible methods inspiring AI resilient architectures. The aim is not only to develop AI algorithms that work well, but also to develop strategies to adapt to the behavior of the users.
In \cite{chacon2025cooperative}, a quantitative approach has been proposed to analyze resilience in agent systems that operate in dynamic and adversarial environments. The authors stated that resilience is not only an individual property, but a capacity to manage in a cooperative through different types of intelligence that can maintain high decision quality and operational stability. 
In order to develop AI resilient systems, generative AI models such as GANs and LLMs have been analyzed \cite{andreoni2024enhancing}. The study stated that generative AI can be used not only to improve human-robot interaction, but also a method of adaptation, prediction, and behavior adjustment. Generative AI can be integrated with reinforcement learning in order to improve the resilience of the system in terms of ability to generate real-time new solutions to respond to unexpected stimuli or behaviors.
In the same approach, another study \cite{umbrico2023mind} address the creation of a social robot capable of a resilient behavior in order to support continuous and adapted human interactions. According to the authors, social robots need to take into account user models (characteristics, social context, and dynamics of interaction) and use a dual-process model operating in different times. The purpose is to promote an empathetic interaction with the user based on verbal and non-verbal cues.
The work Security Considerations in AI-Robotics: A Survey of Current Methods, Challenges, and Opportunities \cite{neupane2024security} aims to provide a structured and comprehensive overview of the main safety issues in AI-powered robotic systems. The authors highlight that the increasing adoption of AI in robotics exposes these systems to a broad range of vulnerabilities, from physical manipulation and cyberattacks to malfunctions caused by corrupted inputs or learning errors. 
The work Disentangling Uncertainty for Safe Social Navigation using Deep Reinforcement Learning \cite{flogel2024disentangling} proposes a deep reinforcement learning-based architecture for the autonomous navigation of social robots in complex, dynamic environments shared with humans. The author emphasizes that social robots must be capable of situational understanding, anticipating others’ intentions, and acting collaboratively—not just avoiding obstacles, but interacting harmoniously with people and moving objects. 
The study \cite{bossens2022resilient} discusses the resilience in social robots designed to assist the elderly. A multimodal architecture has been proposed, in order to recognize when something goes wrong in perception and respond by switching to other ways of communicating or offering support. The paper emphasizes how important it is to design resilience not just from a technical perspective, but with a focus on emotional responses and the ability to adapt to relationships.

\subsection{Resilient AI in Facial Emotion Recognition}
The paper Adaptive Multilayer Perceptual Attention Network for Facial Expression Recognition \cite{liu2022adaptive} introduces AMP-Net, a neural network with multilevel perceptual attention inspired by the human visual system, designed to improve facial expression recognition (FER) under complex real-world conditions such as varying lighting, facial occlusions, and pose variations. 
The study Auto-FERNet: A Facial Expression Recognition Network With Architecture Search  \cite{li2021auto} presents Auto-FERNet, a lightweight neural network specifically tailored for FER, automatically generated using differentiable Neural Architecture Search (NAS) directly on FER datasets. Unlike generic architectures such as VGG or GoogLeNet, which are often too heavy and unsuitable for FER in uncontrolled environments, Auto-FERNet is optimized for this specific task. 
The paper Facial Expression Recognition Robust to Occlusion and to Intra-Similarity Problem Using Relevant Subsampling \cite{kim2023facial} is focuses on two critical challenges: facial occlusions and the high similarity between expressions within the same class. The proposed approach uses a neural network equipped with an attention mechanism and a Spatial Transformer Network (STN) to emphasize the facial regions most relevant to each emotional state. 
Another study Efficient Facial Expression Recognition With Representation Reinforcement Network and Transfer Self-Training for Human–Machine Interaction  \cite{jiang2023efficient} presents an efficient approach to FER in the context of human-machine interaction, combining a Representation Reinforcement Network (RRN) with a Transfer Self-Training (TST) method. 
In \cite{wang2022rails}, authors present RAILS (Robust Adversarial Immune-inspired Learning System), a novel defense framework against adversarial attacks, inspired by the adaptive human immune system. RAILS begins with a balanced population of examples across different classes and a uniform label distribution to foster diversity. 

\subsection{Resilient AI in Natural Language Processing systems}
In \cite{omar2022robust} authors provide a structured overview of robustness in natural language processing (NLP) systems. Although recent advancements in deep learning have significantly improved performance on benchmark datasets, NLP models remain vulnerable to adversarial attacks, revealing fundamental shortcomings in language understanding and limitations in real-world deployment. 
The paper A Survey of Adversarial Defences and Robustness in NLP \cite{goyal2023survey} highlights over the past few years, it has become evident that deep neural networks are susceptible to adversarial perturbations in input data. Numerous studies have introduced effective attack strategies in both computer vision and NLP domains, prompting the development of various defense mechanisms to prevent model failure. 
The work Natural Language Processing Unveiled: Overcoming Challenges, Celebrating Milestones, and Navigating Emerging Research Frontiers \cite{christophernatural} also presents a thorough overview of the evolution of NLP, tracing its development from early rule-based systems to contemporary deep learning approaches. 
The article Improving the Reliability of Deep Neural Networks in NLP: A Review \cite{alshemali2020improving} addresses the vulnerability of deep learning models in NLP to adversarial examples—inputs that have been subtly modified to deceive the model. This fragility represents a critical barrier to the adoption of neural networks in security-sensitive environments. 
The study Exploring Vulnerabilities and Protections in Large Language Models: A Survey \cite{liu2024exploring} focuses on the security vulnerabilities of large language models (LLMs), specifically examining two categories: prompt hacking and adversarial attacks. Prompt hacking is explored through detailed explanations of prompt injection and jailbreaking techniques, analyzing how they work, the risks they pose, and potential countermeasures.
This contribution Measure and Improve Robustness in NLP Models: A Survey \cite{wang2021measure} provides a unified survey on robustness in NLP models, a crucial aspect for the safe application of these systems in real-world settings. 
\section{Conclusion}
This section discusses the results and evidence presented in the review. This paper shows how resilience is a fundamental characteristic of social robots, which, through it, ensure trust in the robot itself—an essential element especially when operating in contexts with elderly people, who often have low trust in these systems.
Resilience is therefore the ability to operate under adverse or stressful conditions, even when degraded or weakened, while maintaining essential operational capabilities.
This review also highlights how new AI systems must be brought to a high level of resilience, in order to ensure a high level of reliability; this is especially essential within the healthcare sector, where these systems support and make decisions regarding treatments and care for patients.
To achieve good levels of resilience in artificial intelligence models, it is crucial to consider various aspects already in the early stages of the development process, particularly during data preparation, through Data Augmentation and the subsequent model evaluation.
A more systematic review of resilience in the fields of social robotics and resilient AI is currently being developed, following the PRISMA 2020 framework \cite{page2021prisma}. In the present article we have highlighted central perspectives drawn from ecology, psychology, engineering, and socio-technical research. The forthcoming review will go beyond this selective synthesis by providing a transparent account of search strategies, screening procedures, and inclusion criteria. We see that work as a natural continuation of the study presented here: while this paper lays out the conceptual and methodological ground, the extended review will supply a comprehensive evidence base that can support both theoretical refinement and practical guidance for designing resilient and socially responsive robotic systems.


\begin{credits}

\subsubsection{\ackname} This work has been partially supported by the Italian PNRR
MUR project PE0000013 - SPOKE 3 ‘RESILIENT AI’, CUP E63C22002150007. It is supported within the framework of the National Recovery and Resilience Plan (NRRP), Mission 4 “Education and Research” – Component 2 “From Research to Business”, with explicit reference to funding provided by the European Union through the NextGenerationEU initiative.

\end{credits}

\printbibliography

\end{document}